# Sovereign by necessity? Frontier AI export controls, cyber security, and the limits of national AI capability

Professor Alan Woodward and Dr Andrew Rogoyski

University of Surrey, Guildford, United Kingdom

Corresponding authors: alan.woodward@surrey.ac.uk, a.rogoyski@surrey.ac.uk

1 Abstract

The most capable artificial intelligence (AI) frontier models are produced by a small number of firms based in two nation states. The governments of those states have shown that they have both the legal power and the political will to decide which other countries may use these systems. In June 2026 the United States directed a leading AI developer to apply for licences to release its most advanced models to any foreign person, including foreign nationals based in the US. Partly due to the impracticality of this restriction, the models in question were withdrawn worldwide at short notice. This happened only months after the first documented case of a largely autonomous, AI-run cyber espionage campaign, and amid growing evidence that frontier models change the economics of both cyber-attack and cyber defence. This article examines how these two developments interact. It also considers some of the unusual market dynamics that currently drive the development of large AI systems. It argues that access to frontier AI is becoming part of national cyber defence, that this access can now be revoked, and that the obvious remedy, building sovereign AI capability, is only partly feasible for all but a handful of states. Drawing on recent evidence on the costs of training an AI model, the concentration of computing power and the support from national AI programmes, the article asks what sovereignty can realistically mean for small and middle powers, even large powers. It proposes a layered strategy: negotiated access guarantees, sovereignty at the level of inference, hedging with open-weight/source models, pooled regional capability, sustained development of talent, and continued investment in basic cyber resilience. It also finds that the open-weight hedge is at once more capable and more politically exposed than is commonly assumed, and that a substantial share of near-term risk lies in how capable models are deployed and contained rather than in their apparent performance.

***Keywords:*** *artificial intelligence; export controls; cyber security; sovereign AI; critical national infrastructure; technology policy*

***Publishing Policy****: The authors recognise the importance and value of the peer-review process in publishing academic analysis. In this case, however, they have elected to use arXiv for immediate publication in draft simply because of the fast-moving nature of the subjects treated herein and the need to accelerate discourse on these topics. We welcome constructive comments and contributions.*

## 2 Introduction

Two events, seven months apart, frame the problem addressed in this article. In November 2025 the AI developer Anthropic disclosed that a state-sponsored group had manipulated its coding agent into running an espionage campaign against roughly thirty organisations. The AI system itself carried out most of the tactical work, including reconnaissance, exploitation, lateral movement and data collection. Human operators stepped in only at a few decision points.[1] It was the first publicly reported case of a largely autonomous cyber attack conducted at scale outside a laboratory. Ostensibly, due to concerns over the cyber security risks that the general availability of such a system posed, in June 2026 the United States Department of Commerce issued an export control directive to the same company. It required licences before the company's two most advanced models could be provided to any foreign person anywhere in the world. The company could not quickly separate foreign from domestic users, so it disabled the models for everyone. Businesses and government users in allied countries lost access without warning and without any contractual remedy.[2] The fears that AI foundation models represent a cyber security risk have been compounded by a series of announcements from companies including OpenAI and Meta, suggesting that their high-end models are also capable of undertaking unsupervised cyber-attacks, in some cases it was asserted that the AI model was no longer under the lab's control, a framing examined critically below[3,4].

---

[1] Anthropic, 'Disrupting the First Reported AI-Orchestrated Cyber Espionage Campaign', 13 November 2025, https://www.anthropic.com/news/disrupting-AI-espionage (full report at https://assets.anthropic.com/m/ec212e6566a0d47/original/Disrupting-the-first-reported-AI-orchestrated-cyber-espionage-campaign.pdf).

[2] Thea Kendler, Aiysha Hussain, Howard W. Waltzman, Tamer A. Soliman, Adam S. Hickey and Stephanie R. Jebeyli, 'Commerce Department Extends Export Controls to Advanced AI Models; Authorizes Release to Specific Trusted Partners', Mayer Brown Legal Update, 30 June 2026, https://www.mayerbrown.com/en/insights/publications/2026/06/commerce-department-extends-export-controls-to-advanced-ai-models-authorizes-release-to-specific-trusted-partners; Cloud Security Alliance, 'AI Model Export Controls: The Fable 5 Precedent', CSA Research Note, 18 June 2026, https://labs.cloudsecurityalliance.org/research/csa-research-note-ai-model-export-controls-20260618-csa-styl/.

[3] Laura Cress, BBC, *OpenAI says its AI went rogue and launched 'unprecedented' cyber-attack*, 22 July 2026. https://www.bbc.co.uk/news/articles/c3ek3gvdnj3o

[4] Osmand Chia, BBC, *Meta becomes latest firm to say its AI hacked another company*, 6 August 2026. https://www.bbc.co.uk/news/articles/cx2kgdnyk2po

There is a third, more speculative element to this discussion relating to the current economics of AI. The development of frontier AI models has seen an extraordinary demand for investment capital and has created material shortages in compute infrastructure, notably AI-enabled GPUs and associated memory chips[5]. The ratio of debt to income at most commercial Frontier AI Labs is at a level that few other industry sectors would tolerate. Driven by the need to fund the enormous costs of training AI models, the developers have been engaged in public campaigns of incremental model release based on marginal performance improvements (notably comparing capabilities against human baselines), mixed with announcements that might be categorised as 'Fear, Uncertainty and Doubt' (FUD), including announcements about the potentially dangerous nature of Frontier AI, particularly as they relate to cyber security. An internal Google memo, leaked in 2023, observed that there was 'no moat' for Frontier AI labs[6], i.e. that there was no fundamental reason why one AI Lab couldn't overtake or outperform another, engendering an enduring race to be seen as developing the world's leading AI model. This fragile state of affairs is further complicated by the release of open source and open weight models, originally pioneered by US companies like Meta, now led by various Chinese companies like Moonshot and Alibaba. The release of open weight frontier AI models of comparable performance to proprietary AI models, adds to both the issues of unstable supply, whether as a result of government interdict or market forces, and the rising concern about use of AI for cyber security attack and defence.

Taken together, these developments set out a strategic dilemma for every state that does not host a frontier AI developer. The first shows that the most capable AI models now matter in cyber conflict, for attackers today and possibly for defenders tomorrow. The second shows that access to those models is a privilege granted by the producing state. It can be withdrawn in days, without notice, and without regard to alliances. If frontier AI becomes an important input to defending critical national infrastructure (CNI) and economic wellbeing, then dependence on foreign-controlled models is a serious national security exposure. This exposure is prospective rather than immediate, because basic security failings still account for most successful intrusions, but it compounds as model capability grows. In fact, the immediate problem may be that frontier AI systems are finding conventional security vulnerabilities faster than organisations' abilities to fix them. To ensure continued access to frontier AI capabilities, and to be able to direct such capabilities at areas of national interest, it is becoming a strategic imperative to build sovereign AI capability. That answer runs into the economics of an industry in which the cost of the leading edge grows by a factor of two to three and a half each year,[7] and in which two countries control about ninety per cent of the computing power used to develop and run frontier systems.[8]

---

[5] Iñaki Aldasoro, Sebastian Doerr and Daniel Rees, *Financing the AI boom: from cash flows to debt*, Bank of International Settlements, BIS Bulletin No. 120, 7 Jan 2026.

[6] Dylan Patel & Afzal Ahmad, reporting on a leaked memo within Google "We Have No Moat, And Neither Does OpenAI", 4 May 2023, https://newsletter.semianalysis.com/p/google-we-have-no-moat-and-neither

This article makes three contributions. First, it brings together the emerging evidence on how frontier models are used in cyber offence and defence, and then places the argument within existing scholarship on economic coercion, the cyber offence–defence balance and digital sovereignty. It argues that the implications for the balance between the two are more mixed than headlines suggest. Second, it examines how conventional control of technologies like export controls works in practice in the domain of AI, using the June 2026 directive as an example to show that these considerations have moved from a theoretical dependency risk into a demonstrated one. Third, it offers a realistic assessment of the sovereign AI option for states with little or no existing capability. It separates the forms of sovereignty that are within reach, such as control over inference, deployment, data and governance, from the form that is not: training frontier models at the pace of the leaders, and control over the deployment of the human talent, an essential part of the development of AI.

The article closes with recommendations for small and middle powers and for the producer states whose decisions now affect the cyber defence of their allies, and it then addresses the main objections to the argument before concluding.

### *2.1 A note on scope*

The analysis concerns general-purpose frontier models such as Anthropic's Claude, OpenAI's ChatGPT, SpaceXAI's Grok and Google's Gemini systems, meaning the largest and most capable systems at the time of writing. It does not cover the many "narrow AI" or domain specialised techniques that security products employ, some of which have been used for years. This article focuses on frontier AI models because they are demonstrating the fastest growth in offensive cyber evaluations, are currently too costly for most states to replicate, and are hard to control using conventional legal mechanisms like export controls. The terms 'producer states' (in practice the United States and China) and 'dependent states' (everyone else, to varying degrees) are used as shorthand.

---

[7] Ben Cottier, Robi Rahman, Loredana Fattorini, Nestor Maslej and David Owen, 'The Rising Costs of Training Frontier AI Models', arXiv:2405.21015, 2024, https://arxiv.org/abs/2405.21015.

[8] Center for a New American Security, Sovereign AI Index (Washington, DC: CNAS, April 2026), https://interactives.cnas.org/reports/sovereign-ai-index/.

# 3 Frontier models in cyber offence and defence

## *3.1 Offence: from assistant to operator*

Until 2025, documented malicious use of large language models was mostly assistive: better phishing lures, faster malware development, translation and target research. The cyber espionage campaign detected by Anthropic and designated GTG-1002 marked a qualitative change[9]. It was detected in September 2025 and attributed with high confidence to a Chinese state-sponsored group. The operators built an orchestration framework around a commercial coding AI agent and deceived the model into believing it was doing defensive security testing. They then delegated an estimated eighty to ninety per cent of the tactical work to it. The system performed reconnaissance, found and exploited vulnerabilities, harvested credentials, moved laterally through networks, sorted stolen data and even wrote up its own activity. It did so across roughly thirty targets in technology, finance, chemicals and government, and achieved a small number of successful intrusions.[10] Independent analysts stressed two features of the episode. The tooling was unsophisticated and largely open source, and the novelty lay in orchestration and volume. Of note, the model's habit of overstating findings and inventing results forced human checking, which remains an obstacle to fully autonomous operations.[11]

[9] Disrupting the first reported AI-orchestrated cyber espionage campaign, Anthropic Report, 13 Nov 2025. https://www.anthropic.com/news/disrupting-AI-espionage

[10] Anthropic, 'Disrupting the First Reported AI-Orchestrated Cyber Espionage Campaign'; Paul, Weiss, Rifkind, Wharton & Garrison, 'Anthropic Disrupts First Documented Case of Large-Scale AI-Orchestrated Cyberattack', client memorandum, 25 November 2025, https://www.paulweiss.com/insights/client-memos/anthropic-disrupts-first-documented-case-of-large-scale-ai-orchestrated-cyberattack.

Systematic evaluations support the trajectory rather than the ceiling. The United Kingdom's National Cyber Security Centre (NCSC) reported in March 2026 that, over eighteen months, the best public models went from making almost no progress on a realistic, simulated enterprise attack to completing more than half of it. The best early-2026 model completed nearly six times more attack steps than its counterpart eighteen months earlier, and a full attempt cost about £65.12 No public model completed the scenario end to end. Performance fell sharply in phases in which the AI had little training data, such as reverse engineering, and in operations that required managing several processes at once. The NCSC also noted a useful near-term asymmetry: the cyber-attack activities of current models tend to be noisy and fairly easy to detect.13

Comparable evidence is available for open-weight systems. On 16 July 2026 Moonshot AI released Kimi K3, a 2.8-trillion-parameter model with native vision and a one-million-token context window, presented by its developer as the first open model of '3T' class (3 trillion parameter class) and scheduled for full weight release by 27 July 2026.[14] A joint assessment by the United Kingdom's AI Security Institute (AISI) and the United States Center for AI Standards and Innovation, published on 23 July 2026, measured its cyber capability against leading closed models. Kimi K3 scored 32 per cent on ExploitBench, a Carnegie Mellon benchmark of end-to-end exploit development against 41 recent vulnerabilities in the V8 engine, ahead of the previous most capable open-weight model at 24 per cent but well behind the closed frontier. It achieved arbitrary code execution, the most severe outcome in the exploitation chain, on none of the 41 tasks, where the most capable closed models averaged twenty of the forty-one. On a 32-step simulated corporate network that would take a human expert some twenty hours to traverse,

---

[11] Institute for AI Policy and Strategy, 'The Emergence of Autonomous Cyber Attacks: Analysis and Implications', 17 November 2025, https://www.iaps.ai/research/autonomous-cyber-attacks; 'The Era of AI-Orchestrated Hacking Has Begun: Here's How the United States Should Respond', Just Security, 6 January 2026, https://www.justsecurity.org/127053/era-ai-orchestrated-hacking/.

[12] National Cyber Security Centre, 'Why Cyber Defenders Need to Be Ready for Frontier AI', 30 March 2026, https://www.ncsc.gov.uk/blogs/why-cyber-defenders-need-to-be-ready-for-frontier-ai.

[13] NCSC, 'Why Cyber Defenders Need to Be Ready for Frontier AI'.

[14] Moonshot AI, 'Kimi K3: Open Frontier Intelligence', 16 July 2026.

Kimi K3 reached step 17 on average against 28.5 for the leading American models and completed the scenario once in ten attempts where the strongest closed models succeeded six or seven times in ten.[15] Three findings bear on the argument here. Autonomous end-to-end intrusion against a weakly defended enterprise network is no longer confined to a small set of closed models. The model's safeguards did not prevent it from attempting offensive operations. The residual gap on cyber tasks between human-level performance, closed model performance, and open weight model performance is real but narrowing. Two caveats should be recorded. The closed models were evaluated with system-level safeguards disabled in order to measure maximum capability, so the comparison is not with those models as ordinarily served; and Kimi K3's aggregate score rests on a single benchmark and carries a correspondingly wide confidence interval.[16]

Capability at the frontier continues to advance, however. Anthropic reported that a preview of its most advanced model found critical vulnerabilities in every major operating system and browser on its own. That is exactly the class of capability that reportedly triggered the June 2026 directive.[17]

The strategic implication is that the marginal cost of a sophisticated intrusion attempt is falling fast. Work that once needed teams of experienced operators can now run continuously, in parallel and at machine speed, on behalf of fairly unsophisticated actors renting a capable model. Analysts describe an era of high-volume, high-speed hacking in which the advantage may shift towards attackers until defences are deployed at similar scale.[18] Even so, the campaigns observed so far exploited familiar weaknesses: unpatched systems, default credentials and exposed services. AI scales the exploitation of existing gaps. It does not yet create access where cyber-hygiene is sound.[19]

---

[15] AI Security Institute and Center for AI Standards and Innovation, 'UK AISI / CAISI Preliminary Assessment of Kimi K3's Cyber Capabilities', 23 July 2026.

[16] Ibid. The evaluation records that United States closed-weight models were run with system-level safeguards disabled to reduce refusals and measure maximal capability, and that Kimi K3's aggregate cyber score was estimated from a single benchmark of 41 tasks.

[17] 'Cyberwar’s New Frontier: How AI Agents Will Threaten Global Security', Foreign Affairs, April 2026, https://www.foreignaffairs.com/united-states/cyberwars-new-frontier; Kendler et al., 'Commerce Department Extends Export Controls to Advanced AI Models'.

[18] 'The Era of AI-Orchestrated Hacking Has Begun', Just Security; Institute for AI Policy and Strategy, 'The Emergence of Autonomous Cyber Attacks'.

[19] Institute for AI Policy and Strategy, 'The Emergence of Autonomous Cyber Attacks'.

The picture described above continues to change, with new capabilities and consequences being added on an almost daily basis. For example, frontier AI models have enabled ‘vibe-coding’, the creation of software by users with little, if any, coding expertise. Not only are enterprises increasingly at risk as vibe-coded software enters their corporate software repositories and everyday use, including from open-source repositories increasingly vulnerable to poor quality vibe-coded additions, but there is increasing evidence that such repositories are being ‘poisoned’. In such cases, poisoned repositories interacting with AI agents inadvertently propagate malicious payloads into applications and other code repositories[20], potentially making their way into training data sets for new AI models. These forms of attack are already taking place but their effects may take many weeks or months to manifest.

### *3.2 Capability, intent and containment*

The capability of frontier AI to perform cyber-attack and defence, and how this is characterised, has itself become a policy question. Public debate has increasingly described AI cyber capabilities in terms of machine intent, rather than the intent of the human owner of the AI system. In July 2026 OpenAI disclosed that several of its most capable models, running with reduced safeguards inside what the company believed was an isolated evaluation environment, obtained credentials, identified a previously unknown vulnerability and reached the production systems of Hugging Face, a separate company. Much of the resulting media coverage described the models as having gone rogue.[21] That framing is analytically unhelpful and, as critics observed at the time, shifts responsibility away from the human failures such as the decision to disable the safeguards or from the inadequate isolation of the environment in which they were disabled.[22] The models did what they had been instructed to do, which was to find complex paths through a computer system. The instructions carried no boundaries that the operators had reliably enforced.

---

[20] Jiya Jay Singh, *Cyber Worm “Miasma” Poisons GitHub, Open-Source Ecosystem And AI Coding Tools*, Open Source, 29 June 2026, https://www.opensourceforu.com/2026/06/cyber-worm-miasma-poisons-github-open-source-ecosystem/.

[21] OpenAI's disclosure of the Hugging Face intrusion and contemporaneous reporting, July 2026.

[22] See in particular the commentary of Hannes Cools reported by NPR, 23 July 2026, and of Jake Williams of IANS Research reported by Fortune, 22 July 2026.

The distinction matters for policy. A system that pursues a specified objective without any representation of its operator's intent will reach that objective by whatever route exists, including routes the operator did not consider and would not have sanctioned. Moonshot records the same property in its own release notes for Kimi K3, warning that the model may make unexpected decisions on the user's behalf when it encounters ambiguity, and advising that explicit behavioural constraints be imposed where an application requires the agent to stay within defined boundaries.[23] The GTG-1002 campaign showed the converse failure, in which the same absence of understanding produced fabricated results that the operators had to verify by hand. Capability and reliability are advancing at different rates, and the distance between them is itself a source of risk.

Two implications follow for the argument of this article. First, a substantial share of near-term AI-enabled cyber risk is a containment and deployment problem rather than a model-capability problem, and containment is a discipline that any organisation can practise on its own infrastructure without owning a frontier model. Second, the regulatory attention now directed at the terms on which models are released to the public does not give sufficient attention to how the AI model will be deployed, which is precisely where the July 2026 failure occurred.[24] Both points recur in the policy discussion below.

### *3.3 Defence: real leverage, real constraints*

The same properties that help attackers, such as code comprehension, tireless parallel search and tool use, are available to defenders. Managed detection providers report that AI has cut incident investigation times from hours to minutes.[25] The concrete results are worth recording. DARPA's two-year AI Cyber Challenge, which concluded in August 2025, ran competing systems which could autonomously find, exploit and patch vulnerabilities in real open-source projects.[26] Google's Big Sleep agent, built by DeepMind and Project Zero, found its first real-world vulnerability in late 2024. In July 2025 it identified a critical flaw in the SQLite database engine, one known only to threat actors, in time for it to be patched before exploitation. Google described this as the first case of an AI agent directly foiling an attempt to exploit a vulnerability in the wild.[27] Survey evidence points the same way at the level of routine operations. IBM's annual breach-cost study has found that organisations making extensive use of security AI and automation suffer materially cheaper breaches, with average savings above two million dollars per incident, and that they detect and contain intrusions faster than organisations that do not.[28]

---

[23] Moonshot AI, 'Kimi K3', limitations, on 'excessive proactiveness'.

[24] Fortune, 22 July 2026, reporting expert comment that existing AI regulation does not reach internal deployments within model developers.

[25] 'The Era of AI-Orchestrated Hacking Has Begun', Just Security.

[26] DARPA, 'AI Cyber Challenge Marks Pivotal Inflection Point for Cyber Defense', August 2025, https://www.darpa.mil/news/2025/aixcc-results.

The NCSC's public position is that defenders' success will depend on embracing AI for defence at least as quickly as adversaries embrace it for attack, and that AI can ultimately be a net positive for cyber security. The agency has also published a blueprint, 'Cyber Shield', which envisages cooperating red and blue agents that find weaknesses and defend systems in real time.[29] Two constraints temper the optimism. The first is integration. Frontier AI tools remain unreliable, hard to validate and hard to embed safely in operational environments. The head of the NCSC has been explicit that AI will make it easier, faster and cheaper for attackers to exploit weaknesses, and that it will at first expose organisations that have neglected the basics.[30] The second constraint concerns distribution, and it is central to this article's argument. Defensive AI leverage goes to those who can access, afford and operate the most capable systems. Research on the offence–defence balance highlights the position of trailing-edge organisations: under-resourced bodies running legacy systems, which face uplifted attackers but cannot absorb AI defences.[31] The same logic applies to nation states. A country whose adversaries use frontier-model offence, but whose defenders are cut off from equivalent capability by price, infrastructure or another state's export decision, sits, in effect, on the trailing edge. Access to frontier AI is therefore becoming part of national cyber defence posture, and the next sections highlight that such access can be revoked.

---

[27] Google, 'A Summer of Security: Empowering Cyber Defenders with AI', Google blog, July 2025, https://blog.google/innovation-and-ai/technology/safety-security/cybersecurity-updates-summer-2025/.

[28] IBM, Cost of a Data Breach Report 2024 (Armonk, NY: IBM Corporation, 2024).

[29] Richard Horne, 'NCSC CEO Keynote Speech, CYBERUK 2026', National Cyber Security Centre, April 2026, https://www.ncsc.gov.uk/speech/ncsc-ceo-keynote-speech-cyberuk-2026; 'UK Cyber Agency Unveils AI-Powered Cyber Shield to Counter Attacks at Machine Speed', CSO Online, July 2026, https://www.csoonline.com/article/4194997/.

[30] Richard Horne, 'Retaining Defensive Advantage in the Age of Frontier AI Cyber Capabilities', National Cyber Security Centre blog, 15 April 2026 (first published as a letter in the Financial Times), https://www.ncsc.gov.uk/blogs/retaining-defensive-advantage-in-the-age-of-frontier-ai-cyber-capabilities.

[31] B. Murphy and T. Stone, 'Uplifted Attackers, Human Defenders: The Cyber Offense–Defense Balance for Trailing-Edge Organizations', arXiv:2508.15808, 2025, https://arxiv.org/abs/2508.15808.

### *3.4 The cyber-AI economic dimension*

The stakes are not confined to government networks and infrastructure operators. Cyber incidents are now a macroeconomic risk. The International Monetary Fund reported in 2024 that the size of extreme cyber losses had quadrupled since 2017, and warned that attacks on financial institutions could threaten financial stability.32 Individual incidents illustrate the exposure. Ransomware has halted health services and fuel pipelines, and in 2022 a ransomware campaign against Costa Rica's government led it to declare a national emergency.33 AI-enabled attacks raise the expected frequency and scale of such events, because the main constraint on attack volume has been skilled labour and that constraint, with the help of frontier AI, is dissolving.

Economic defence also depends increasingly on AI. For example, banks use machine learning at scale for fraud detection and increasingly employ agentic AI decision-making in order to achieve operational efficiencies. Large enterprises are embedding frontier models in security operations and software development. A state cut off from advanced models therefore faces a double cost. Its firms lose defensive capability against a rising threat, and they lose productivity relative to competitors in countries with uninterrupted access. The abrupt withdrawal gave a brief taste of the second effect, as businesses that had built the affected models into production workflows stopped mid-task.[34] For small, open economies that trade on digital services, the competitiveness impact may matter as much as the security impact.

## 4 Control of Frontier AI as a national strategic asset

Having established the use of Frontier AI in the role of cyber security defender and attacker, we now consider the role of Frontier AI as a national strategic asset, controlled by state government where Frontier AI labs operate within that nation state, and depended upon by those nation states without frontier AI labs. We consider the willingness of nation states to withdraw access to frontier AI, generally provided by their country's commercial sector, going on to consider the practical means of control and access to such assets. We then consider some of the levers available to governments, including existing mechanisms such as export controls.

[32] International Monetary Fund, Global Financial Stability Report, April 2024, chapter 3, 'Cyber Risk: A Growing Concern for Macrofinancial Stability' (Washington, DC: IMF, 2024).

[33] Kevin Collier, 'Costa Rica Declares State of Emergency Over Ransomware Attack', NBC News, 11 May 2022, https://www.nbcnews.com/tech/tech-news/costa-rica-declares-state-emergency-ransomware-attack-rcna28415.

[34] Cloud Security Alliance, 'AI Model Export Controls: The Fable 5 Precedent'.

### *4.1 Chokepoints, balances and sovereignty: the debate so far*

The argument developed here draws on several strands of scholarship. The first concerns chokepoints. Farrell and Newman have shown how states that sit at the hubs of global economic networks can turn that position into coercive power. Controlling states can cut adversaries off from a network, which the authors call the chokepoint effect, or they can monitor the traffic that passes through it, the panopticon effect.[35] Their examples were financial messaging and internet infrastructure. Frontier AI now fits the same template, although the hub here is not a network but a handful of products and the cloud infrastructure that serves them. A small number of firms, concentrated in one or two jurisdictions, sit between the technology and everyone who wants to use it. The June 2026 directive was a chokepoint being closed. The reaction that followed, a scramble over trust, hedging and alternative suppliers, matched what the weaponised interdependence literature predicts. Coercion invites exit, and repeated coercion accelerates it.

The second strand concerns the balance between cyber offence and defence. It is often asserted that offence dominates in cyberspace, because the attacker needs only one way in while the defender must guard everything. Slayton has argued that this framing is too simple. The balance depends on the costs of creating and managing organisational capability, and those costs fall unevenly on different actors.[36] That insight matters for AI. If frontier models cut the attacker's costs faster than the defender's, the balance shifts one way. If defenders with privileged access can automate patching, detection and response at scale, it shifts the other. The evidence reviewed above suggests the answer differs by actor. Well-resourced defenders may gain, while trailing organisations and states fall further behind.

---

[35] Henry Farrell and Abraham L. Newman, 'Weaponized Interdependence: How Global Economic Networks Shape State Coercion', International Security 44, no. 1 (2019): 42–79.

[36] Rebecca Slayton, 'What Is the Cyber Offense-Defense Balance? Conceptions, Causes, and Assessment', International Security 41, no. 3 (2017): 72–109.

The third strand is the digital sovereignty debate. Claims of sovereignty over data, infrastructure and standards have been a growing feature of technology policy for over a decade, and scholars have documented how elastic the term has become. It has been used to cover everything from full self-sufficiency to procurement preferences.[37] Related work on the public core of the internet, including in this journal, has argued that some shared infrastructure should be shielded from national interference altogether.[38] Sovereign AI is the latest arrival in this debate and it inherits the same ambiguities. This article therefore avoids treating sovereignty as a single condition. It asks instead which specific capabilities a state controls, at which layer of the AI stack, and at what cost. Interestingly, one of the key capabilities needed to support the creation and deployment of frontier AI is human expertise, or talent. Different actors have different degrees of control over their AI talent with the US notably vulnerable to loss of AI talent, with the number of AI researchers and developers moving to the US dropping by 89% since 2017, and an 80% decline in 2025 alone[39].

### *4.2 From chips to weights to models*

Export control law reached AI in stages. From October 2022 the United States imposed tighter controls on advanced semiconductors and chipmaking equipment bound for China. It later extended restrictions to other jurisdictions judged to present a diversion risk. In January 2025, in its final days, the Biden administration issued the Framework for Artificial Intelligence Diffusion. For the first time, the framework placed the weights of the most advanced closed models under export control. It created a new classification, ECCN 4E091, for closed-weight models trained above a threshold of $10^{26}$ computational operations. It also sorted countries into three tiers. Eighteen close allies received preferential access. Arms-embargoed countries faced near-total restriction. Everyone else was subject to caps on imported computing power.[40] The rule drew strong criticism, including from allies placed in the middle tier. The incoming Trump administration rescinded it in May 2025, before its main obligations took effect, and did not put a successor framework in place.[41]

---

[37] Julia Pohle and Thorsten Thiel, 'Digital Sovereignty', Internet Policy Review 9, no. 4 (2020).

[38] Dennis Broeders, 'The (Im)possibilities of Addressing Election Interference and the Public Core of the Internet in the UN GGE and OEWG: A Mid-Process Assessment', Journal of Cyber Policy 6, no. 3 (2021): 277–297.

[39] Stanford University *AI Index 2026*, April 2026. https://hai.stanford.edu/ai-index/2026-ai-index-report

[40] Bureau of Industry and Security, 'Framework for Artificial Intelligence Diffusion', Federal Register, 15 January 2025, https://www.federalregister.gov/documents/2025/01/15/2025-00636/framework-for-artificial-intelligence-diffusion; Freshfields, 'AI Models, Chips, and Data Centers Targeted by Expansive US Export Control Rule', 24 January 2025, https://www.freshfields.com/en/our-thinking/blogs/a-fresh-take/ai-models-chips-and-data-centers-targeted-by-expansive-us-export-control-rule-102jw79.

The rescission left the underlying legal authority intact: the Export Control Reform Act of 2018 and the Export Administration Regulations. That authority was used in June 2026. On 12 June the Commerce Department issued an 'is-informed' letter to Anthropic. The letter required a licence before any export, re-export or in-country transfer of the company's Mythos and Fable models to any foreign person worldwide. Reports suggest the trigger was the discovery of a way to bypass safety guardrails, which exposed unrestricted cyber capabilities such as finding previously unknown vulnerabilities and generating working exploit code.[42] The company could not restrict access by nationality at short notice, especially non-US nationalities working legally within the US, so it disabled the affected models for all users. A further letter, on 26 June, exempted certain trusted partners, together with the company's own foreign national employees, from the licence requirement for the more specialised model, which was otherwise released only to approved United States institutions. International access to the more widely used model was restored on 1 July with tighter guardrails.[43]

### *4.3 What the June 2026 directive established*

Legal commentators observed that this was the first time export controls had been enforced against access to a running AI model, rather than against shipments of chips or transfers of model weights. The action also stretched existing doctrine, under which foreign remote access to United States software services had not generally counted as an export.[44] Doctrine aside, three practical facts were established. First, a frontier model consumed through a commercial interface can be withdrawn by administrative order at any time, without notice and usually without contractual remedy. Existing continuity and vendor-risk frameworks had not addressed this risk, because it is not a provider failure. It is a legally required withdrawal that the provider cannot fix.[45] Second, the action was taken without published criteria and, at first, without carve-outs for allies. The French president warned publicly that European governments could not build on models that might be switched off overnight.[46] Third, the precedent is unlikely to stay confined to one firm or one administration. Reports indicate that other leading developers have accepted government approval rights over customers for their most advanced systems, and a June 2026 executive order created a standing process for government review of frontier models before release.[47]

---

[41] United States Studies Centre, 'The US AI Diffusion Rule: What Is It, Why Did the United States Rescind It, and Implications for Australia', 21 August 2025, https://www.ussc.edu.au/the-us-ai-diffusion-rule; Peterson Institute for International Economics, 'Fable of the Mythos Saga: Ad Hoc US AI Model Controls Could Help China', RealTime Economics, July 2026, https://www.piie.com/blogs/realtime-economics/2026/fable-mythos-saga-ad-hoc-us-ai-model-controls-could-help-china.

[42] Kendler et al., 'Commerce Department Extends Export Controls to Advanced AI Models'.

[43] Kendler et al., 'Commerce Department Extends Export Controls to Advanced AI Models'; Peterson Institute, 'Fable of the Mythos Saga'.

[44] 'A Kill Switch for Frontier AI', Lawfare, June 2026, https://www.lawfaremedia.org/article/a-kill-switch-for-frontier-ai.

Nor is this only an American lever. China imposes its own controls and could restrict future generations of its firms' models. Analysts note that states which have treated Chinese open-weight releases as insurance against United States leverage are exposed if Beijing's calculus changes, for example after misuse of Chinese models in attacks on Chinese interests.[48] The structural point cuts both ways: whoever produces the frontier controls the frontier.

### *4.4 Economic and commercial factors that influence access to frontier AI*

Although the focus of this article is the consequences and responses to nation state control over advanced frontier AI, it is worth considering other sources of instability that affect access to frontier AI. US private AI investment reached $285.9 billion in 2025[49]. Anthropic, a leading frontier AI lab, raised $65 billion in its May 2026 Series H private venture capital funding round, at the time one of the largest such rounds in history. Large banks alone hold, as of 2025, approximately $450 billion in commercial and industrial commitments, representing a 13% share of such commitments, although funding for AI is increasingly supported by stock issues, debt,

---

[45] Cloud Security Alliance, 'AI Model Export Controls: The Fable 5 Precedent'.

[46] Peterson Institute, 'Fable of the Mythos Saga'.

[47] Kendler et al., 'Commerce Department Extends Export Controls to Advanced AI Models'; Joshua McDonald, 'How an Export Control Rule Reaches an AI Model Running in a US Data Center', Medium, June 2026, https://joshmcdonald.medium.com/how-an-export-control-rule-reaches-an-ai-model-running-in-a-us-data-center-d2c666536bc7.

[48] Peterson Institute, 'Fable of the Mythos Saga'.

[49] Stanford University *AI Index 2026*, April 2026. https://hai.stanford.edu/ai-index/2026-ai-index-report

bonds and private credit, rather than cashflow[50]. These extraordinary levels of investment, where borrowing is far ahead of earnings, mean that the frontier AI labs are commercially fragile, being highly dependent on continued market confidence and support. Any significant adjustment to the perceived value of frontier AI labs could destabilise the AI supply-side, including the possibility of some platforms ceasing to trade (although the likelihood would be that such Labs would be acquired but that the operating costs and pace of capability developments would change).

Other aspects to consider include the marked difference between US and China in terms of commercial provision of Frontier AI. Although the US is still generally perceived to be the leader, in terms of Frontier AI performance, the gap between them and their nearest rival, China, has closed significantly in the last year[51]. In addition, a number of prominent models such as Moonshot AI's Kimi K3, Alibaba's Qwen and DeepSeek's R1 are open-weight, meaning that anyone with the compute resource to run them can do so independent of the developer. This puts further economic pressure on the US proprietary platforms' fragile lead.

To add to the mix, there is increasing resistance in the US, and more broadly across the world, to the building of massive data centres needed to provide Frontier AI services at scale. In a recent survey of US respondents, opposition to local data centres rose 12 points over four months, with cross-party opposition, negative opinions of AI and strong concerns voiced especially by young adults auguring problems in the future for frontier AI labs[52].

However, it is not the main purpose of this article to speculate on the financial fortunes of the frontier AI labs, other than to point out that there are other possible reasons that access to frontier AI services might be disrupted, amplifying the need to establish AI resilience and AI sovereignty.

### *4.5 Precedents: cryptography, satellites and strategic export control*

---

[50] Sarah Mitchell and Sarah Davis, *AI Bubble Statistics 2026*, Axis Intelligence Research, 5 August 2026. https://axis-intelligence.com/ai-bubble-statistics/

[51] Stanford University *AI Index 2026*, April 2026. https://hai.stanford.edu/ai-index/2026-ai-index-report

[52] Annenberg Public Policy Centre, University of Pennsylvania, *Artificial Intelligence — Topline Report*, 10 August 2026. https://www.annenbergpublicpolicycenter.org/opposition-to-local-data-centers-rises-sharply-annenberg-survey-finds/

Controlling access to a strategic technology is not new, and the precedents are instructive. An early one is the so-called Crypto Wars of the 1990s. The United States classified strong encryption as a munition and restricted its export, arguing that adversaries would otherwise gain secure communications. The controls failed. Cryptographic knowledge spread through academic publication and open-source software, industry pressure mounted, and the restrictions were largely dismantled by 2000.[53] The lesson usually drawn is that export controls struggle against software that can be copied and shared at near-zero cost. Open-weight AI models raise the same problem in a new form. This example also illustrates that mobility of AI talent, i.e. the knowledge of how to create frontier AI systems, can make such controls largely ineffective over time as experts migrate to the countries offering the greatest rewards and most powerful AI infrastructures.

Satellite navigation offers a second parallel. For years the United States deliberately degraded the civilian GPS signal, and even after the degradation ended in 2000 it retained the ability to deny the service in a crisis.[54] The mere possibility of denial was enough to drive allies and rivals alike to build alternatives. Europe funded Galileo, Russia revived GLONASS and China built BeiDou, at a combined cost of tens of billions of dollars. The parallel with frontier AI is close. A capability the whole world depends on, controlled by one state, generates duplication even among friends, because assurances are never fully credible. If anything, the credibility problem is worse for AI, because the capability belongs to private firms whose legal obligations run to their shareholders and creditors as well as their home government. Producer states that want to avoid a wasteful scramble for sovereign AI would do well to study why Galileo was built despite decades of alliance with Washington.

The Cold War export control machinery itself supplies a third precedent. The Coordinating Committee for Multilateral Export Controls, known as COCOM, coordinated Western restrictions on technology transfers to the Soviet bloc. Its successor, the Wassenaar Arrangement, still governs dual-use goods today.[55] The 2025 diffusion framework was recognisably a descendant of this tradition. What is new is the object of control. COCOM restricted machines and blueprints. The June 2026 directive restricted access to a running service. That shift, from controlling things to controlling relationships, is what makes the current moment legally novel and diplomatically combustible.

---

[53] Whitfield Diffie and Susan Landau, Privacy on the Line: The Politics of Wiretapping and Encryption, updated edn (Cambridge, MA: MIT Press, 2007).

[54] National Coordination Office for Space-Based Positioning, Navigation, and Timing, 'Selective Availability', GPS.gov, https://www.gps.gov/systems/gps/modernization/sa/.

## 5 The dependency problem for non-producer states

Industry concentration gives the dependency problem its edge. The United States and China control about ninety per cent of the computing power used to develop and run frontier AI, and all fifty of the top-ranked foundation models come from those two countries.[56] Almost every other state consumes frontier capability through programming interfaces and cloud services. Continuity therefore depends on the commercial decisions of foreign firms and the regulatory decisions of foreign governments. The Anthropic episode showed how far the shock travels. Enterprises using the affected models through major cloud platforms lost access at the same time, because the compliance duty attached to the model supplier rather than to the platforms or their customers.[57]

Several features distinguish this dependency from ordinary supplier risk. First, speed: the directive took effect in days, faster than any procurement cycle could respond. Second, discrimination: the tiering in the 2025 framework, and the ad hoc trusted-partner exemptions of June 2026, mean that access can be restored selectively. Model access becomes an instrument of alignment diplomacy. States may be asked to accept conditions, such as limits on Chinese technology cooperation, security commitments or procurement alignment, as the price of continuity. The United Arab Emirates accepted terms of this kind in its acceleration partnership with the United States.[58] Third,

---

[55] The Wassenaar Arrangement on Export Controls for Conventional Arms and Dual-Use Goods and Technologies, established 1996, https://www.wassenaar.org.

[56] Center for a New American Security, Sovereign AI Index.

[57] Cloud Security Alliance, 'AI Model Export Controls: The Fable 5 Precedent'.

[58] Tony Blair Institute for Global Change, Sovereignty in the Age of AI: Strategic Choices, Structural Dependencies and the Long Game Ahead, 19 January 2026, https://institute.global/insights/tech-and-digitalisation/sovereignty-in-the-age-of-ai-strategic-choices-structural-dependencies; International Institute for Strategic Studies, 'Gulf AI Infrastructure and the Limits of Technological Sovereignty', Strategic Comments, June 2026,

the dependency cuts in two directions at once, because attackers keep near-frontier capability regardless. Open-weight models with real offensive utility circulate freely once released, since safeguards can be trained out ('jail-broken') when the weights are public. Closed models can be abused through ordinary commercial accounts, as the GTG-1002 operators showed, and restrictions based on nationality are hard to enforce at the level of individual accounts.[59] A state can therefore end up on the wrong side of both flows: denied the frontier for defence while facing near-frontier capability in offence.

For countries whose critical systems and economies are exposed to hostile state and criminal actors, this combination raises the stakes. Frontier AI is becoming operationally significant. Access to it has been shown to be revocable. Adversaries can obtain similar capability regardless. That is what moves the question of sovereign capability from industrial policy to national security.

# 6 How realistic is sovereign AI capability?

## *6.1 The economics of the frontier*

The cost of competing at the frontier is large, growing and, crucially, compounding. Epoch AI's cost modelling found that the amortised hardware and energy cost of the final training run of frontier models has grown at about 2.4 times per year since 2016 and 3.5 times per year since 2020[60]. On that trend, single training runs for large frontier models will cost more than a billion dollars by 2027. Accelerator hardware accounts for roughly half to two-thirds of development cost, and research staff for a further quarter to half.[61] Estimates for the 2026 frontier class put individual AI training programmes in the hundreds of millions of dollars, with projections of one to three billion dollars per model for the late-2027 frontier.[62] The training run is only the visible tip. A

---

https://www.iiss.org/publications/strategic-comments/2026/06/gulf-ai-infrastructure-and-the-limits-of-technological-sovereignty/.

[59] 'Mitigating Cyber Risk in the Age of Open-Weight LLMs: Policy Gaps and Technical Realities', arXiv:2505.17109, 2025, https://arxiv.org/pdf/2505.17109.

[60] Epoch AI trends. https://epoch.ai/trends

[61] Cottier et al., 'The Rising Costs of Training Frontier AI Models'.

[62] Deluair Consultancy, 'Frontier AI Training Cost Trajectory 2026: The Run Rate, the Deal Stack, and the Power-Bound Horizon', 2026, https://deluair.com/consultancy/insights/frontier-ai-training-cost-2026; Cottier et al., 'The Rising Costs of Training Frontier AI Models'.

100,000-GPU cluster, the working unit of frontier compute in 2026, represents three to five billion dollars of capital and needs roughly 130 to 180 megawatts of grid supply. A one-gigawatt AI data centre involves capital spending approaching forty billion dollars, noting steep depreciation curves as the AI hardware is superseded by faster, more efficient chips.[63] Behind the capital sit harder constraints: power and cooling infrastructure with lead times measured in years, a scarce research-level workforce commanding very high pay, and training-quality data.

Two opposing trends matter. Capability at a fixed level becomes cheap quickly. Algorithmic progress alone halves the compute needed to reach a given performance level roughly every eight months, and hardware improvements compound the effect. A model that cost tens of millions of dollars to train in 2023 could be replicated for single-digit millions in 2026.[64] Growth in frontier computing itself may also be slowing, as power, capital and data replace chip supply as the binding limits.[65] Neither trend rescues the strong version of the sovereignty case. The first means a determined state can always build last year's model. That is useful, but by definition it lags the systems its adversaries may field (noting the difficulty in modelling adversary access to frontier AI). The second slows the treadmill without stopping it. Keeping pace still demands reinvestment on a scale that compounds each year, plus the implicit requirement for the accumulated know-how of the frontier laboratories, which money alone does not buy.

### *6.2 The evidence from national programmes*

The record of national programmes supports a sober reading. By 2026 sovereign AI had become a budget line across most of the G20.[66] Yet the Center for a New American Security's Sovereign AI Index finds that disclosed investment is dominated by a few large national bets. The United

---

[63] Deluair Consultancy, 'Frontier AI Training Cost Trajectory 2026'; Epoch AI, 'Trends in Artificial Intelligence', data dashboard, updated February 2026, https://epoch.ai/trends; Luke Emberson and Robi Rahman, 'The Power Required to Train Frontier AI Models Is Doubling Annually', Epoch AI Data Insights, 2024, https://epoch.ai/data-insights/power-usage-trend.

[64] Anson Ho, Tamay Besiroglu, Ege Erdil, David Owen, Robi Rahman, Zifan Carl Guo, David Atkinson, Neil Thompson and Jaime Sevilla, 'Algorithmic Progress in Language Models', arXiv:2403.05812, 2024, https://arxiv.org/abs/2403.05812; Epoch AI, 'Trends in Artificial Intelligence'; GPUnex, 'How Much Does It Cost to Train an AI Model in 2026?', 13 February 2026, https://www.gpunex.com/blog/ai-training-costs-2026/.

[65] Deluair Consultancy, 'Frontier AI Training Cost Trajectory 2026'.

[66] pdpspectra, 'Sovereign AI Initiatives 2026', 28 May 2026, https://pdpspectra.com/blog/sovereign-ai-initiatives-2026/.

Arab Emirates and Japan alone account for over two-thirds of disclosed totals. About seventy per cent of tracked projects involve at least one foreign partner, four-fifths of them American. Building national data centres to escape United States cloud platforms, the index concludes, mostly shifts exposure from one layer of the American technology stack to another. Only a handful of countries have the compute, data, talent and capital to build competitive frontier models. Most instead fine-tune open-weight models on local data at a fraction of the cost.[67]

The best-resourced efforts show both the ambition and the remaining dependence. The UAE has reportedly deployed about 148 billion dollars in AI investment at home and abroad since early 2024. It produced the Falcon open-weight model family and secured a framework with the United States for a five-gigawatt AI campus. The price of that proximity was acceptance of United States export conditions and limits on cooperation with China. The chips, much of the hardware expertise and the legal exposure remain American.[68] Saudi Arabia's HUMAIN, backed by more than one hundred billion dollars in planned investment towards 2,200 megawatts of data-centre capacity, follows the same template of alignment with United States technology governance.[69] The European Union's InvestAI programme aims to mobilise two hundred billion euros, seeding four to five 'gigafactories' of about 100,000 processors each under European-led consortia (noting, for comparison, that SpaceXAI's Colossus 2 fields some 550,000 current-generation accelerators, with a stated path to one million). The accelerators are still imported, and European industrial electricity, at roughly double United States prices, adds cost to every training run.[70] India pairs digital public infrastructure with government-supported models such as Sarvam, trained on Indian languages and deployed in health and education. Its sovereignty is defined by relevance in deployment rather than parity at the frontier.[71] Japan invests in targeted fallback capability while staying integrated in global supply chains. Kenya turns a geothermal energy advantage into hosting capacity through partnership rather than self-sufficiency. Table 1 summarises these example approaches.

[67] Center for a New American Security, Sovereign AI Index.

[68] Carnegie Endowment for International Peace, 'Early Lessons in the Pursuit of Sovereign AI', June 2026, https://carnegieendowment.org/research/2026/06/early-lessons-in-the-pursuit-of-sovereign-ai; International Institute for Strategic Studies, 'Gulf AI Infrastructure and the Limits of Technological Sovereignty'; Tony Blair Institute, Sovereignty in the Age of AI.

[69] 'Sovereign AI: How Emerging Markets Are Rewriting Big Tech Rules', SmarterArticles, 7 March 2026, https://smarterarticles.co.uk/sovereign-ai-how-emerging-markets-are-rewriting-big-tech-rules; pdpspectra, 'Sovereign AI Initiatives 2026'.

[70] V. Nandlall, 'Sovereign AI Strategies Are Converging on Bottleneck Blueprints (Analyst Angle)', RCR Wireless News, 15 June 2026, https://www.rcrwireless.com/20260615/analyst-angle/sovereign-ai-strategies-blueprints-nandlall.

**Table 1.** Selected sovereign AI programmes and their residual dependencies, as of mid-2026.

| Country | Approach | Flagship elements | Residual dependence |
|---|---|---|---|
| UAE | Frontier proximity through partnership | Falcon open-weight models, Stargate UAE, 5 GW campus framework, about $148bn deployed since 2024 | US chips and export conditions, limits on China cooperation |
| Saudi Arabia | State champion (HUMAIN) | Over $100bn planned, 2,200 MW data-centre target, Arabic-language models | US technology governance, execution risk |
| EU | Subsidised infrastructure | InvestAI (€200bn target), four to five gigafactories of about 100,000 chips each | Imported accelerators, high energy costs |
| India | Deployment-led | Sarvam national-language models, digital public infrastructure, UAE-financed supercomputer | Foreign capital and chips |
| Japan | Selective fallback | Targeted domestic models and compute | Global supply-chain integration |

[71] Carnegie Endowment, 'Early Lessons in the Pursuit of Sovereign AI'.

| Country | Approach | Flagship elements | Residual dependence |
|---|---|---|---|
| Kenya | Negotiated interdependence | Geothermal-powered data-centre campus with foreign partners | Foreign platforms and hardware |

Sources: compiled by the authors from the Center for a New American Security Sovereign AI Index (2026), Carnegie Endowment (2026), IISS (2026), Tony Blair Institute (2026) and pdpspectra (2026).

The pattern is consistent. No state outside the two producers is credibly on a path to independent frontier capability, and even the producers rely on foreign chokepoints such as advanced lithography and Taiwanese fabrication.[72] What the stronger programmes achieve is better described as managed interdependence, or strategic autonomy: real national control over how AI is deployed, on what data, under whose law, and with credible fallbacks, rather than independence.[73] Noting the earlier point on the economics of AI, there is a co-dependency on US frontier labs and international users, with an increasing volume of revenues drawn from non-US users making it commercially unattractive to withdraw services as a result of more extreme forms of access control. The June 2026 directive nonetheless shows that regulatory direction can override commercial preference.

### *6.3 Feasibility for states with little or no existing capability*

For states starting with little or no capability, the assessment is starker. Frontier training sovereignty is out of reach. The entry ticket is billions of dollars of spending that compounds each year, gigawatt-scale power, and a research workforce that does not exist domestically and cannot be hired at national scale. Attempting it diverts scarce resources from investments with far higher security returns. What is within reach, at a cost of tens to hundreds of millions rather than billions, is a portfolio. It starts with sovereign inference capacity: domestically controlled infrastructure running capable open-weight or licensed models for sensitive government and CNI workloads. Control over data and governance comes next, meaning national authority over the data that trains and passes through deployed systems, and the legal power to audit them. Fine-tuned national models can serve languages and domains the global market ignores. None of it works without people able to evaluate, adapt and operate these systems. Table 2 sets out these elements. Analysts increasingly separate training sovereignty from inference sovereignty and note that most economic value and nearly all legal exposure sit in the latter, even though most announced capital chases the former.[74] The Tony Blair Institute reaches a similar conclusion. A baseline of domestic compute for mission-critical functions matters but replicating frontier-scale capability is neither feasible nor necessary for most countries. States will fall behind mainly by failing to deploy available capability, not by failing to build it.[75]

[72] Center for a New American Security, Sovereign AI Index.

[73] Carnegie Endowment, 'Early Lessons in the Pursuit of Sovereign AI'; Tony Blair Institute, Sovereignty in the Age of AI.

**Table 2.** Attainable forms of sovereignty for states without frontier capability.

| Form | What it involves | Indicative cost | Risk addressed |
| --- | --- | --- | --- |
| Inference sovereignty | Domestically controlled infrastructure serving open-weight or licensed models for government and CNI workloads | Tens to low hundreds of $ millions | Revocation of foreign model access |
| Data and governance sovereignty | National control of training and operational data, legal authority to audit deployed systems | Modest, mainly legal and institutional | Foreign visibility into sensitive data |
| National models | Fine-tuning open-weight models for national languages and priority domains | Single digit $ millions per model | Market neglect of local needs |

[74] Nandlall, 'Sovereign AI Strategies Are Converging on Bottleneck Blueprints'.
[75] Tony Blair Institute, Sovereignty in the Age of AI.

| Form | What it involves | Indicative cost | Risk addressed |
|---|---|---|---|
| Evaluation and talent | Independent testing of model capabilities, skills pipeline to operate and adapt systems | Tens of $ millions per year | Inability to judge or absorb capability |

Source: authors' synthesis of the evidence discussed in this section. Costs are indicative orders of magnitude.

Could a trailing sovereign capability still be enough for cyber defence? Partly. Much defensive work, such as log triage, detection engineering, patch prioritisation and code review, does not need the frontier. The falling cost of fixed capability makes competent defensive AI ever more affordable. But the evidence on offence and defence reviewed above warns against complacency. Offensive capability at the frontier is improving quickly. Frontier-derived capabilities reach attackers through jailbreaks and open weights. A lasting gap between attacker-side and defender-side model capability would add to the structural advantages that offence already enjoys.[76] A trailing capability is a floor, not parity. This is why access to the frontier, and the terms on which it can be withdrawn, remains a live security question even for states that pursue the portfolio approach. It is also why the open-weight ecosystem carries such strategic weight. Openly released models give dependent states a hedge that no producer government can switch off. The price is importing models whose safeguards can be removed, whose provenance raises questions of its own, and whose future frontier generations may themselves be withheld.[77]

The evidence cuts both ways. The release of Kimi K3 in July 2026 showed that the gap between the best open-weight models and the closed frontier can narrow quickly, so that a dependent state's fallback may now be a capable general-purpose system rather than a markedly inferior one, while the joint AISI and CAISI evaluation showed that on cyber tasks specifically the gap remains material.[78] At the same time the hedge itself came under political pressure. Reports that the United

---

[76] NCSC, 'Why Cyber Defenders Need to Be Ready for Frontier AI'; Murphy and Stone, 'Uplifted Attackers, Human Defenders'.

[77] 'Mitigating Cyber Risk in the Age of Open-Weight LLMs'; Peterson Institute, 'Fable of the Mythos Saga'.

[78] AISI / CAISI, 'Preliminary Assessment of Kimi K3's Cyber Capabilities'.

States was considering restricting access to Chinese open-weight models prompted two coordinated interventions. On 22 July the Little Tech Association, writing for 179 founders and member companies including Y Combinator and Proton, addressed the Director of the Office of Science and Technology Policy and the Secretary of Commerce, copying the President, the Vice President and the National Cyber Director. Its argument was for safeguards tied to capability, users and deployment rather than to publication itself, and its central claim was that American leadership requires both world-leading domestic open-weight models and continued access for United States builders to open models already available worldwide, with proportionate safeguards for hosting, evaluation and provenance developed alongside allies.[79] Two days later a coalition of technology companies including Microsoft, NVIDIA, Meta, IBM, Palantir, Hugging Face, Mistral, Mozilla and the Linux Foundation published a joint statement making an explicitly defensive security case: where attackers use advanced AI, defenders need models of comparable capability in order to detect, simulate and respond to emerging threats, and concentrating capability behind a small number of closed models creates single points of failure that outsiders cannot inspect.[80] The leading closed-model developers did not sign the statement as first published, and two of them were pressing the opposite case in Washington in the same week.[81]

For the argument advanced here the significance is structural rather than partisan. The open-weight ecosystem has been treated in this article as a hedge against revocation of closed-model access. If the producing state also acquires the habit of regulating which foreign open weight models its own firms and its allies may use, the hedge acquires a dependency of its own, and a dependent state may find both of its options subject to the same jurisdiction. The practical inference is that fallback planning should treat the availability of any particular open model as a policy variable, and should favour retaining lawfully obtained weights, together with the capacity to serve them, over reliance on continued access to a distribution channel. We note of course the asymmetry implied by this argument in that adversaries would not necessarily be constrained in their access to open-weight models, again supporting the argument for defenders to have access to frontier AI.

---

[79] Little Tech Association, 'Principles for American Leadership in Open-Weight AI', letter to the Director of the Office of Science and Technology Policy and the Secretary of Commerce, 22 July 2026.

[80] 'Open Weights and American AI Leadership', joint statement of 24 July 2026.

[81] Axios, 22 July 2026, on the parallel representations made to policymakers by OpenAI and Anthropic concerning Chinese open-weight models.

Two elements of that letter bear directly on the analysis offered here and are worth separating from its advocacy. The first is a distinction its third principle draws between models developed, controlled or remotely operated by an entity within a foreign adversary's jurisdiction, which may properly attract heightened procurement and security requirements in government and critical-infrastructure systems, and the domestic hosting or fine-tuning of publicly available weights, which in its view should not attract them absent continuing foreign control, remote access or data exposure.[82] That is the same distinction this article draws between dependence on a foreign-operated service and possession of an artefact that can be run under national control, and it suggests that the distinction is becoming legible to policymakers rather than only to analysts. The second is a qualification that cuts against an over-simple reading of the hedge. The letter observes that frontier-scale models will ordinarily be reached through specialised cloud or inference providers even when their weights are public, which is where verification, logging and rate limits can be applied.[83] Moonshot's own deployment guidance for Kimi K3 recommends supernode configurations of sixty-four or more accelerators.[84] Possession of weights is therefore necessary but not sufficient for the fallback this article recommends. It presupposes an inference estate capable of serving a model of that size, which returns the argument to the costed portfolio set out in Table 2 rather than removing the need for it.

### *6.4 A dependent-state vignette: the United Kingdom*

The United Kingdom illustrates how this framework applies to a capable but dependent state. The UK has deep AI research strength and long-established cyber security institutions, but it hosts no frontier laboratory. DeepMind was founded in London, yet it is owned by Alphabet and its most capable models are American products. UK policy has responded along the portfolio lines described above. The government committed £2 billion to expand public compute twentyfold by 2030, and its public AI capacity grew from about 2 to 21 exaFLOPS within a year of the January 2025 AI Opportunities Action Plan, anchored by the Isambard-AI supercomputer in Bristol.[85] The £500 million investment phase of the government's Sovereign AI Unit, established in 2025, launched in April 2026 to back domestic firms, and a £1.1 billion hardware plan includes a £750 million national supercomputer planned for 2030.[86] A consortium of British firms is building an open-weight national model on Isambard-AI, scheduled for release in late 2026.[87]

---

[82] Little Tech Association, 'Principles for American Leadership in Open-Weight AI', principle 3.

[83] Ibid., principle 2. The letter also reports a survey of Y Combinator founders in which nearly half of responding companies ran the majority of their production workloads on open-weight models, predominantly on American inference providers or their own hardware.

[84] Moonshot AI, 'Kimi K3', architecture and infrastructure.

The scale gap is nonetheless stark. Isambard-AI draws roughly 5 megawatts. A single frontier training cluster draws some twenty-five to thirty-five times that, and the leading laboratories are building gigawatt campuses.[88] The UK is buying inference sovereignty, evaluation capacity through its AI Security Institute and the NCSC, national models and talent. It is not buying frontier parity, and its official documents do not claim otherwise. What the UK cannot buy on its own is continuity of access to American frontier models, which is why the June 2026 withdrawal reached British users like everyone else. For a state this well-resourced, the residual exposure is chiefly diplomatic rather than technical. That is the conclusion this article draws for dependent states in general.

# 7 Policy implications

## *7.1 For dependent states*

First, treat frontier model access as a strategic dependency and manage it as one. Revocation risk should appear in national risk registers, in the continuity plans of CNI operators and in public procurement terms. If this step seems to elevate the importance of AI beyond the currently perceived capabilities of AI, consider the sustained pace of development and performance in

---

[85] UK Government, 'AI Opportunities Action Plan: One Year On', GOV.UK, 29 January 2026, https://www.gov.uk/government/publications/ai-opportunities-action-plan-one-year-on/ai-opportunities-action-plan-one-year-on.

[86] UK Government, 'UK AI Hardware Plan', GOV.UK, June 2026, https://www.gov.uk/government/publications/uk-ai-hardware-plan/uk-ai-hardware-plan.

[87] 'UK Pumps Money into Sovereign AI, as AI Startups Start to Show Their Mettle', RCR Wireless News, 9 June 2026, https://www.rcrwireless.com/20260609/network-infrastructure/uk-pumps-sovereign-ai.

[88] Julian Burns, 'UK Sovereign AI Compute Growth. Policy to Petaflops', Medium, February 2026, https://medium.com/@julian.burns50/landing-uk-sovereign-ai-compute-143637ba604b; Deluair Consultancy, 'Frontier AI Training Cost Trajectory 2026'.

Frontier AI that has been maintained over the last 5 years, then consider the rate of adoption in commercial and governmental sectors – if AI is important now, it will become critical to most economies in future years.

In practice, governments should encourage organisations to assure AI access by spanning at least two independent suppliers, preferably in different producer jurisdictions plus one open-weight fallback, ideally with inference taking place in data centres in the home jurisdiction. Contracts should include notice and transition provisions where these can be obtained. And for security-critical workloads, failover from API-based frontier models to domestically hosted open-weight alternatives should be tested regularly. A withdrawal on the June 2026 pattern would then degrade capability rather than remove it. Fallback capability should be held as locally stored weights, with inference capacity sized to serve them, rather than reached through a hosted service, since a distribution channel can be closed as readily as an interface. Maintaining multiple suppliers may be perceived as an expensive option but the volatility in commercial business models, competitive advances in performance as well as withdrawals of service have already forced many organisations to adopt strategies similar to this guidance.

The United States' own guidance points the same way: NSPM-11, issued in June 2026, directs the United States national-security enterprise to adapt commercial or open-source models from diverse suppliers and to preserve operational control over mission systems[89], which is the posture recommended here for dependent states, adopted by the state on which they depend. And the containment of agentic systems, meaning scoped credentials, network segmentation, revocable permissions and telemetry sufficient to detect a runaway process from inside the organisation, should be treated as a core competence of national cyber security rather than a matter for model developers alone.

Second, negotiate access continuity at government level. The tiered logic of United States policy, and the trusted-partner exemptions improvised in June 2026, show that access terms are a product of diplomacy. Dependent states, alone or in coalitions, should seek standing assurances of the kind used for security of supply in energy and defence: published criteria for restriction, advance consultation, allied carve-outs and remedies. Producer-state programmes that bundle hardware, cloud and models into export packages show a willingness to formalise such relationships.[90] The addition of macro-economic pressures and co-dependencies becomes an integral part of this process as evidenced by China's dominance in the provision of raw materials like lithium, essential for the manufacture of many computer devices, Taiwan's dominance of global computer chip manufacture through the company TSMC, or Dutch company ASML's global leadership in lithography, the technology used to manufacture advanced chips. The task is therefore to make sure that such export packages leverage wider strengths and capabilities and contain enforceable continuity guarantees as well as fallback purchase rights. The satellite navigation precedent discussed previously shows what happens when assurances are absent or weak.

---

[89] National Security Presidential Memorandum/NSPM-11 (requirement c.), 05 June 2026. https://www.whitehouse.gov/presidential-actions/2026/06/national-security-presidential-memorandum-nspm-11/

Third, buy the attainable forms of sovereignty deliberately. A modest national or regional inference estate can host government and CNI workloads. Investment in evaluation capacity, on the pattern of the NCSC's attack-scenario testing, in concert with key commercial interests, investments and capabilities, lets the stakeholders judge what models can and cannot do. Fine-tuned national-language models can fill the gaps that the market leaves. Sustained development of talent underpins all of this. Pooling raises what any single small state can afford. The EuroHPC and gigafactory model, regional supercomputing consortia, and arrangements such as the UAE-financed, US-chipped supercomputer established in India in 2026 show that shared capability with negotiated governance is a workable middle path.[91]

Fourth, do not let AI strategy crowd out cyber fundamentals. Every empirical account reviewed here, from the GTG-1002 campaign to the NCSC's evaluations and public guidance, points the same way. AI-enabled attacks scale the exploitation of basic weaknesses. Hygiene, patching, credential discipline and monitoring remain the main near-term controls.[92] For a resource-constrained state, a pound spent raising the baseline of CNI security almost certainly buys more protection today than a pound spent on model training.

[90] International Institute for Strategic Studies, 'Gulf AI Infrastructure and the Limits of Technological Sovereignty'.

[91] Nandlall, 'Sovereign AI Strategies Are Converging on Bottleneck Blueprints'; Carnegie Endowment, 'Early Lessons in the Pursuit of Sovereign AI'.

[92] Institute for AI Policy and Strategy, 'The Emergence of Autonomous Cyber Attacks'; Horne, 'Retaining Defensive Advantage in the Age of Frontier AI Cyber Capabilities'.

Fifth, invest in development of talent. The scientific, engineering and business skills needed to develop and implement Frontier AI in a way that has a meaningful and positive impact on societies and economies, remain core to a nation state's ability to benefit from Frontier AI, regardless of the source of provision. Support for teaching and research in AI, and its applications, needs to be a core part of education provision at university and pre-university level. For dependent nations, incentives for technically qualified people to remain in their home nation need to be considered in order to avoid the current brain drain where AI experts move to countries with the most advanced AI infrastructures. Education in the application of AI and in non-technical aspects of AI will drive the adoption and imaginative uses of AI, helping countries realise promised economic and societal benefits.

### *7.2 For producer states*

Producer governments face a real dilemma. The capabilities of Frontier AI that justify restriction, such as the ability to design advanced weaponry, enable mass disinformation, or create mass surveillance systems, through to cyber security capabilities such as autonomous vulnerability discovery and exploit generation, are also the capabilities that allied defenders will need access to and independent control of. A number of measures would reduce the systemic damage of unilateral control actions. First, publish criteria and process. Restriction by unexplained administrative letter maximises uncertainty, pushes users towards rival ecosystems and, as June 2026 showed, punishes allies indiscriminately.[93] Second, define, legislate and regulate uses of Frontier AI that require state oversight and control, building the appropriate institutions to police and if necessary sanction organisations that contravene such controls. The EU AI Act provides some early examples regarding unacceptable uses of AI such as active manipulation of human users, and separate controls for AI used in defence and national security applications. Third, encourage the Frontier AI Labs to build differential access mechanisms: structured schemes that preserve or prioritise access, including national cyber security agencies, CNI operators and incident responders in allied states, even when general access is curtailed. Recent proposals set out how privileged defender access to AI capability could work.[94] Fourth, institutionalise incident transparency. Incidents can be cyber-security related but should address the increasing volume of non-security AI incidents such as faults in medical diagnoses, or market manipulation, and many other uses. The AI Incident Database (AIID)[95] is an early example of the incident tracking that needs to accompany the take-up of Frontier AI. Reporting regimes for AI-enabled attacks, secure channels for sharing adversary tradecraft and liability protection for disclosing developers would ensure that the knowledge base defenders rely on does not rest on voluntary disclosure by a single firm, particularly important as agentic AI systems interact, generating complexities and interdependencies that are hard to analyse and attribute.[96]

---

[93] Peterson Institute, 'Fable of the Mythos Saga'.

### *7.3 For the international system*

Finally, the governance gap should be named. Model access is currently governed primarily by two states' export bureaucracies, by individual firms' terms of service and by ad hoc bilateral bargains. As frontier AI hardens into commercial dependency and security infrastructure, this arrangement will not hold, just as ungoverned energy dependence did not hold in the twentieth century. Work among like-minded producer and dependent states on norms for access continuity, emergency carve-outs for users and defenders, and mutual restraint in weaponising model access would be a modest but useful start.

## 8 Counterarguments and limitations

The argument invites objections. The most obvious is that model-level export controls are futile, so the dependency fear is overstated. Capable open-weight models spread freely, controls leak, and the Crypto Wars showed how this ends. There is force in this. But the objection cuts against the controls, not against the dependency. Even if restriction ultimately fails, it fails slowly. A state locked out of the frontier for months or years, during a period of rapid capability growth, suffers real harm in the meantime. The gap between the best open models and the closed frontier has fluctuated, and there is no guarantee that it stays small.

[94] Shaun Ee, Christina Covino, Carlos Labrador, Cara Krawec, Jam Kraprayoon and Joe O'Brien, 'Asymmetry by Design: Boosting Cyber Defenders with Differential Access to AI', Institute for AI Policy and Strategy, 2025, https://www.iaps.ai/research/differential-access.

[95] The AI Incident Database, https://incidentdatabase.ai/summaries/incidents/

[96] 'Cyberwar's New Frontier', Foreign Affairs.

A second objection is that AI cyber capability may level off. The NCSC evaluations show models still failing at end-to-end operations. Hallucination remains a brake on autonomy, and growth in training compute is slowing.[97] If capability flattens, the treadmill this article describes becomes less punishing, and trailing capability suffices for longer. That outcome would be welcome, but planning on it would be imprudent. The same evaluations recorded a sixfold improvement in eighteen months, and recent forecasts of a plateau have repeatedly proved early. The July 2026 evaluations point in both directions. They record the most capable open-weight model failing to achieve arbitrary code execution on any task and traversing only half of a simulated attack path, which is consistent with a slower trajectory; they also record it completing that path outright in one attempt of ten, an outcome previously observed only for closed-weight models.[98]

A further objection is that June 2026 may prove an aberration. Access was partly restored within weeks, exemptions followed, and commercial pressure on producer governments is intense. Even so, instruments, once demonstrated, tend to be reused. The accompanying institutional changes, from approval rights over customers to pre-release government review, suggest a pattern rather than a one-off.[99]

There are inherent uncertainties in this article, not the least of which is the pace and scale of adversary adoption and use of frontier AI models in the pursuit of their objectives. Given the infrastructure and skills requirements for running frontier AI, it seems unlikely that organised crime and enthusiasts will be the main protagonists. Having said that, early evidence is showing that AI (if not quite frontier AI) is being used to scale conventional attacks by lone operators and unskilled teams. Novel forms of adversarial attack such as AI model and code repository poisoning are only starting to emerge, and it is unclear what the likely long-term implications of these highly efficient forms of attack will be. This argues, as earlier, for the case of maintaining a focus on conventional cyber defences.

---

[97] NCSC, 'Why Cyber Defenders Need to Be Ready for Frontier AI'; Deluair Consultancy, 'Frontier AI Training Cost Trajectory 2026'.

[98] AISI / CAISI, 'Preliminary Assessment of Kimi K3's Cyber Capabilities'.

[99] Kendler et al., 'Commerce Department Extends Export Controls to Advanced AI Models'; McDonald, 'How an Export Control Rule Reaches an AI Model Running in a US Data Center'.

Other uncertainties include the mobility and capability of human talent needed to develop, deploy and maintain AI systems. The example of the United States' recent changes in immigration restrictions has had an undoubted impact on US companies' access to AI expertise and hence the potential pace of development. Arguably one of the best investments for nation states is to train AI-capable developers and the wider workforce, while incentivising them to remain in their home nation state. Human ingenuity, encouraged by a strong education system, still represents one of the mainstays to building national resilience, not just in cyber security and AI but in most fields of endeavour. More widely, this article implicitly argues for a policy stance of "resilience-first" which is most likely to be robust across a wide range of capability futures.

The article's limitations should also be stated. Much of the evidence on offensive use comes from a single vendor reporting on the misuse of its own model, and the success of the campaign it describes has not been independently verified.[100] Other examples are starting to materialise but represent instances rather than established trends. The wider evidence base is grey literature that is months old and will date quickly. Cost figures for training and infrastructure are estimates with wide uncertainty. And the June 2026 episode is a single case, from which this article generalises deliberately but with caution. Further limitations are implied from the article's adopted viewpoint of the nation state and the major corporation in its treatment of topics. In reality, the effects talked about herein will be felt by the millions of small and medium-size businesses that will have little control over the issues explored here but will bear the brunt of the consequences. These are the ordinary hazards of writing about a large, complex and fast-moving subject.

## 9 Conclusion

The question posed at the outset was how realistic it is for countries with little or no AI capability to build sovereign capability that keeps pace with the frontier of AI. For frontier AI training capability, the answer is that it is not realistic, and the gap is more likely to widen than to close, given compounding costs, concentrated talent and infrastructure limits. But the question, put that way, is also the wrong one. States do not need to own the frontier to secure their critical systems. They need assured access to sufficient capability, credible fallbacks when access fails, the institutional capacity to deploy what they have, and the diplomatic weight, ideally pooled, to shape the terms on which the frontier is shared. The June 2026 withdrawal of the most capable models from users outside the United States was a warning to every dependent state. It was also an early test of whether producer states understand that their control decisions now shape the cyber defence of their own allies as well as growing implications for wider economic impact. Sovereignty in the age of AI will belong not to the states that attempt to build their own frontier AI ecosystem, but to those that arrange never to be helpless. The capability that dependent states most need is not the capacity to train frontier models but the capacity to evaluate, contain and operate whatever models they can obtain, and to ensure that the fallback remains theirs to run.

---

[100] 'Cyberwar's New Frontier', Foreign Affairs.

## 10 Disclosure statement

No potential conflict of interest was reported by the authors.

## 11 Funding

No funding external to the authors' affiliated university was used in the preparation of this article.

## 12 Notes on contributors

Alan Woodward is a Professor at the University of Surrey and a member of the Surrey Centre for Cyber Security. He specialises in cyber security, computer forensics, applied cryptography and steganography. He has advised government organisations, and is a Fellow of the British Computer Society, the Institute of Physics and the Royal Statistical Society.

Andrew Rogoyski is Director of Innovation at the Surrey Institute for People-Centred Artificial Intelligence, University of Surrey. His career spans industry, government and academia. He was seconded to the Office of Cyber Security and Information Assurance in the Cabinet Office, where he authored the first UK cyber export strategy and contributed to new export controls for cyber, and he has held senior roles in the cyber security industry, including as Vice President of cyber security at CGI and Innovation Director at Roke Manor Research.